\documentclass[9pt,conference]{IEEEtran}
\IEEEoverridecommandlockouts
\usepackage{cite}
\usepackage{amsmath,amssymb,amsfonts}
\usepackage{algorithmic}
\usepackage{graphicx}
\usepackage{textcomp}
\usepackage{xcolor}
\usepackage{subcaption}

\def\BibTeX{{\rm B\kern-.05em{\sc i\kern-.025em b}\kern-.08em
    T\kern-.1667em\lower.7ex\hbox{E}\kern-.125emX}}
\begin{document}

\title{Intermittent Inference with Nonuniformly Compressed Multi-Exit Neural Network for Energy Harvesting Powered Devices}

\author{Yawen Wu$^{\dagger}$, Zhepeng Wang$^{\dagger}$, Zhenge Jia$^{\dagger}$, Yiyu Shi$^{\ddagger}$, and Jingtong Hu$^{\dagger}$ \\
$^{\dagger}$Department of Electrical and Computer Engineering, University of Pittsburgh, USA \\
$^{\ddagger}$Department of Computer Science and Engineering, University of Notre Dame, USA \\
Email: yawen.wu@pitt.edu, zhepeng.wang@pitt.edu, zhenge.jia@pitt.edu, yshi4@nd.edu, jthu@pitt.edu\vspace{-0pt}}

\maketitle

\begin{abstract}

This work aims to enable persistent, event-driven sensing and decision capabilities for energy-harvesting (EH)-powered devices by deploying lightweight DNNs onto EH-powered devices. However, harvested energy is usually weak and unpredictable and even lightweight DNNs take multiple power cycles to finish one inference. To eliminate the indefinite long wait to accumulate energy for one inference and to optimize the accuracy, we developed a power trace-aware and exit-guided network compression algorithm to compress and deploy multi-exit neural networks to EH-powered microcontrollers (MCUs) and select exits during execution according to available energy. The experimental results show superior accuracy and latency compared with state-of-the-art techniques. 
\end{abstract}
\begin{IEEEkeywords}
Energy harvesting, intermittent inference, neural network compression
\end{IEEEkeywords}

\section{Introduction}

The maturation of energy harvesting (EH) technology and the recent emergence of intermittent computing, which stores harvested energy in energy storage and supports an episode of program execution during each power cycle, creates the opportunity to build sophisticated battery-less energy-neutral sensors.
EH technology can scavenge energy from the ambient environment, such as solar power \cite{fraternali2018scaling, zhang2016solar, luo2019spoton}, kinetic power \cite{huang2016battery,ju2018power}, and thermal gradient \cite{mitcheson2010energy, wu2018prototyping}.
One of the most promising applications of such sensors is to build persistent, event-driven IoT systems in which the main device (e.g. a battery-draining processing system) can remain dormant, with near-zero power consumption, until awakened by an EH-powered sensor, which monitors events of interest constantly with harvested energy. To realize this capability, the EH-powered sensor has to frequently make decisions locally with sensor data, as it is prohibitive to send the raw data to other devices and offload the computation to them.

Deep neural networks (DNNs) can effectively extract features from noisy input data. However, they are usually computationally expensive. Typical neural networks have tens of millions of weights and use billions of operations to finish one inference. Even a small DNN (e.g. MobileNetV2~\cite{sandler2018mobilenetv2}) has over a million weights and millions of operations. 
However, microcontrollers (MCUs) are constrained in resources. Typical MCUs have limited storage (e.g. Flash or FRAM) size (several or tens of KB) and run in low frequency (several or tens of MHz). Directly deploying DNN to MCU is infeasible because the model size exceeds the storage capacity. Even if the DNN model can fit into the limited storage, the time to finish one inference is still too long (tens or hundreds of seconds).

DNN inference on intermittently-powered devices remains largely unexplored. Existing work \cite{gobieski2019intelligence} made the first step to implement DNNs on an intermittently powered MCU. 
However, multiple power cycles are needed to finish one inference in most cases. Since the harvested power is usually weak and unpredictable, the latency to obtain the final inference result can be indefinitely long. 
Recently, the multi-exit network with classifiers in shallower layers is proposed \cite{teerapittayanon2016branchynet, huang2017multi}. They are very promising for EH-powered devices with limited energy budget because they can reduce the inference energy cost and latency by exiting from early-exits while maintaining the accuracy.   

However, to achieve efficient inference with multi-exit networks on EH-powered devices, 
the first challenge is how to fit the multi-exit network onto MCUs while keeping a high accuracy of each exit. 
Simply compressing the network with existing network compression approaches \cite{gobieski2019intelligence} does not work well since they only consider the accuracy of the final exit. For a multi-exit network, only considering the final exit during compression will significantly degrade the accuracy of early-exits. 
Unfortunately, the EH-powered system often chooses early-exits in shallower layers to generate the result with the limited energy budget, which results in low accuracy. 
Therefore, how to compress the network considering the accuracy and energy cost of each exit remains a problem. It becomes more complicated when the power source is considered. Powered by dynamic EH, the chances that each exit is selected are different depending on both the power condition and the accuracy/energy cost of each exit after compression.
To maximize the average accuracy of all the events, the compression algorithm has to take the power condition and accuracy/energy cost of multiple exits into consideration. 
Maximizing the average accuracy across all the events is equivalent to maximizing the number of interesting events that are correctly processed in a fixed amount of harvested energy, which is important for EH-powered devices.

The second challenge is how to select the exit for each event during runtime to achieve a high average accuracy in the long-term. The exit needs to be selected based on the available EH energy and the difficulty of each input. Two sequential decisions need to be made. First, when an event happens, simply selecting the exit with the highest accuracy that current energy can support can result in low average accuracy in the long run. This is because even if current EH efficiency is high, it can be low in the future. Instead of using up all the available energy for one inference to achieve high accuracy, a better strategy is to reserve some energy for the future events. Otherwise the following events will have low accuracy or even be missed because of insufficient energy. Second, the inference difficulty of each input needs to be considered. The difficulty is only known at an exit by inspecting the entropy of current result. If the confidence is low at the selected exit, a second decision needs to be made on whether an incremental inference is needed to propagate the input to a following exit for a higher accuracy. To make these two decisions, the EH condition and the difficulty of current event need to be considered. 

To address these two challenges, we propose a two-phase approach to automatically compress multi-exit neural networks before deployment and conduct runtime exit selection. 
In the first phase, we aim to compress the multi-exit network to fit it onto MCUs and achieve high average accuracy of all events. First, we will consider typical EH power traces and event distribution, which determine the probability of selecting each exit. Priority will be given to the exits which have higher probability of being selected. Since the probability of selecting each exit will change after we compress the network due to the change of computation complexity for each exit, we develop a reinforcement-learning (RL) based approach to automatically search the best pruning rate, bitwidth of weights and activations in all the layers to maximize the average accuracy. 

In the second phase, we aim to maximize the average accuracy for all the events during runtime. We employ Q-learning~\cite{watkins1992q} to learn the best exit under different EH energy conditions. To select the exit for an event, we use the current available energy level and the charging efficiency as the state, and use all exits as the actions the learning method can take. Q-learning is lightweight as it uses a lookup table (LUT) to select actions. The learning process only involves updating the LUT. To decide whether to conduct incremental inference, we use the confidence of the result at the selected exit and current available energy as the state. The action is a binary decision, representing to continue the inference or to output the current result.

In summary, the main contributions of the paper include:
\begin{itemize}
	\item \textbf{Intermittent inference model.} We propose an intermittent and incremental inference model to guarantee an inference result before power failure occurs. Waiting for the next power cycle is not needed while further refining is still possible. 
	\item \textbf{Power trace-aware compression.} We develop a power trace -aware and multiple exits-guided compression technique to compress multi-exit networks to fit onto MCUs while maximizing the average inference accuracy.
	\item \textbf{Runtime adaptation.} We propose an online exit selection method to select the exit for each event considering the EH condition and difficulty of each input.
\end{itemize}

Experimental results show that the proposed techniques improve the number of correctly processed events per energy unit
by 3.6x over \cite{gobieski2019intelligence}, a state-of-the-art (SOTA) intermittent inference framework. It also outperforms \cite{fedorov2019sparse}, a NAS framework to generate networks for MCUs, by 18.9x. The latency of all the processed events is improved by 7.8x and 10.2x over these two approaches, respectively.

\section{Event-Triggered Intermittent Inference}\label{sec:inferencemodel}

\begin{figure}[!htb]
	\centering 
	\includegraphics[width=\columnwidth]{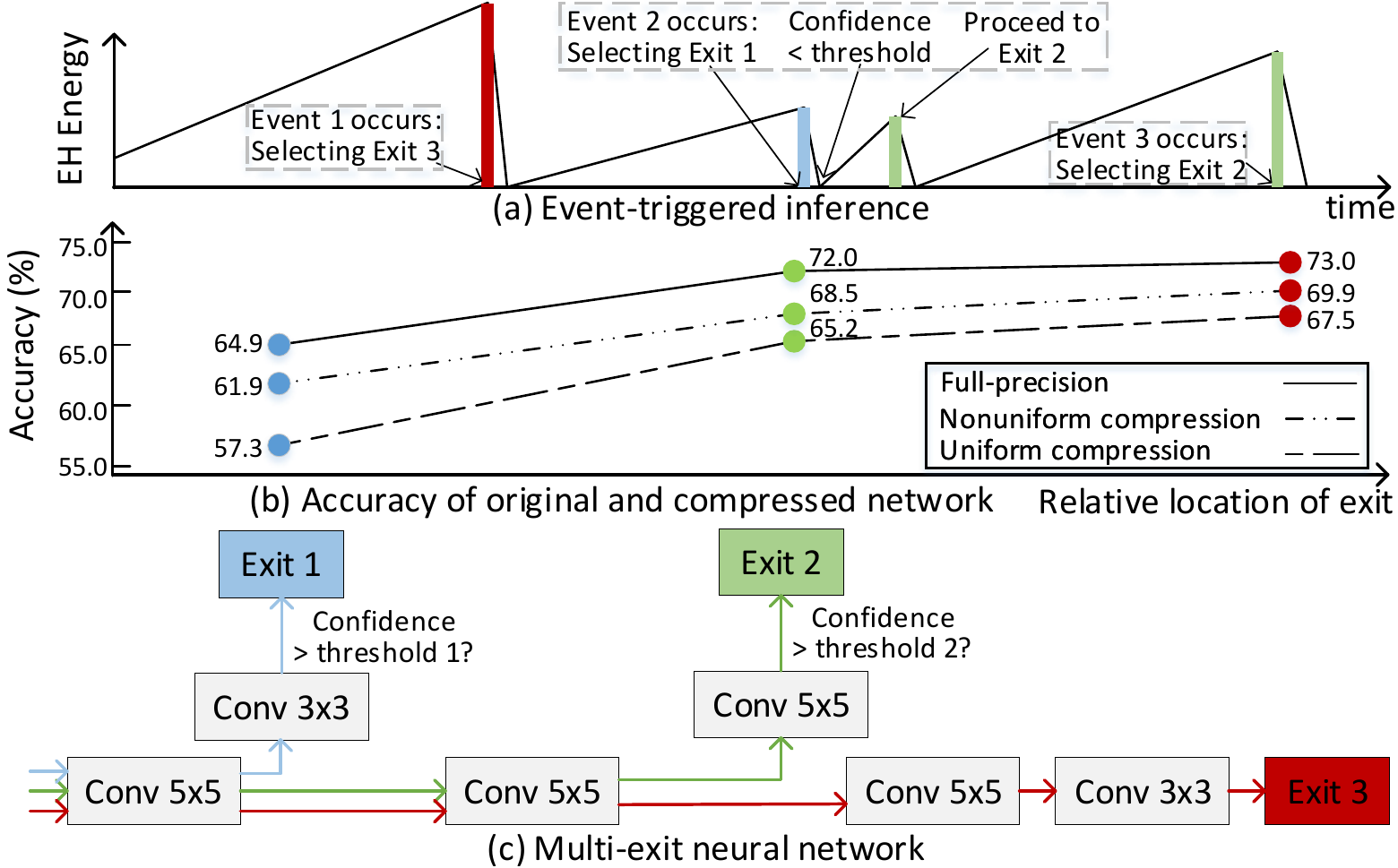}
	\caption{Intermittent execution model with multi-exits and benefits of nonuniform compression.}
	\label{fig:earlyexit}
\end{figure}

In the existing state-of-the-art deployment of DNNs on EH-powered devices \cite{gobieski2019intelligence}, when the power is not sufficient to finish the entire forward-pass, the system is forced to pause during the inference process and wait until enough energy is harvested. However, the unpredictable EH process can result in indefinite waiting time to harvest sufficient energy, by which time the event may become obsolete.
To solve this problem, we employ networks with multi-exits \cite{huang2017multi}. As shown in Figure~\ref{fig:earlyexit}(b)(c), this simple network has 3 exits, and each exit has a different accuracy and energy cost on CIFAR-10. As shown in Figure~\ref{fig:earlyexit}(a), when an event triggers the inference, an exit will be selected according to the available energy and the energy cost of each exit. In this example, when Event 1 occurs, the stored energy is sufficient to support the inference to Exit 3, which is selected as the exit. However, when Event 2 occurs, the energy can only support the inference to Exit 1. At each exit, the confidence of the result is measured by the entropy. If the confidence is higher than a threshold, the inference exits from this point. Otherwise, when more energy is available, an incremental inference will be made to proceed to the following exit for higher accuracy. In this example, since the confidence of Event 2 in Exit 1 is below the threshold, an incremental inference is conducted to proceed to Exit 2.
This process alleviates the indefinitely long waiting time problem and an inference result with confidence can be obtained during each power cycle. 

\textbf{Metric} We use local inference to filter sensor readings from events so that only the interesting events are used to wake up the main device. Our figure of merit is the number of interesting events that are correctly processed in a fixed amount of harvested energy. We denote it as \emph{IEpmJ}, or Interesting Events per milliJoule. Maximizing \emph{IEpmJ} is equivalent to maximizing the average accuracy of all events:
\setlength{\abovedisplayskip}{0pt}
\setlength{\belowdisplayskip}{0pt}
\setlength{\abovedisplayshortskip}{0pt}
\setlength{\belowdisplayshortskip}{0pt}
\begin{equation}\label{eq:metric} \small
IEpmJ=\frac{N_{correct}}{E_{total}}=\frac{\sum_{j=1}^{N_1}Acc_{j}+\sum_{j=1}^{N_2} 0}{E_{total}}=\frac{N}{E_{total}}(\frac{1}{N}\sum_{j=1}^{N}Acc_{j})
\end{equation}
$N_{correct}$ is the number of correctly processed events. $N=N_1+N_2$ is the number of all the events in which $N_1$ events are processed by inference and $N_2$ events are missed due to insufficient energy. 
$N_{correct}$ is a subset of $N_1$ and $N_{correct}=\sum_{j=1}^{N_1}Acc_{j}$.
$Acc_{j}\in\{0,1\}$ where $Acc_{j}=1$ represents event $j$ is correctly processed and $Acc_{j}=0$ otherwise.
Since $N$ and $E_{total}$ are constants determined by the EH environment,
maximizing \emph{IEpmJ} is equivalent to maximizing the average accuracy of all $N$ events, which is the number of correctly processed events over all the events $\frac{1}{N}\sum_{j=1}^{N}Acc_{j}$.

To deploy this inference model, the multi-exit network needs to be compressed to fit onto resource-constrained MCUs. The compression approach will be introduced in Section \ref{sec:compress}.
\section{Power Trace-Aware, Exit-Guided Network Compression}\label{sec:compress}

In this section, we will develop an EH powered trace-aware and exit-guided network compression algorithm. It aims to fit the multi-exit network onto MCUs and maximize the average accuracy by allocating layer-wise pruning rate and quantization bitwidth.
Existing compressing algorithms, which uniformly compress network, will significantly degrade the accuracy of exits in shallow layers as shown in Figure \ref{fig:earlyexit}(b).
Different from existing algorithms that only consider the accuracy of the final classifier, this approach takes the accuracy of all exits into consideration and conducts nonuniform compression. As shown in Figure~\ref{fig:earlyexit}(b), if we take a non-uniform approach, which compresses less in the shallow layers and more in the deep layers, the accuracy drop for all exits will be small. What is more, some exits will be chosen more often than the others under a given power trace and event distribution. Thus, we will prioritize the accuracy of these exits during the compression process. In this way, we can improve the average inference accuracy across all events.

\begin{figure}[!htb]
	\centering
	\includegraphics[width=\columnwidth]{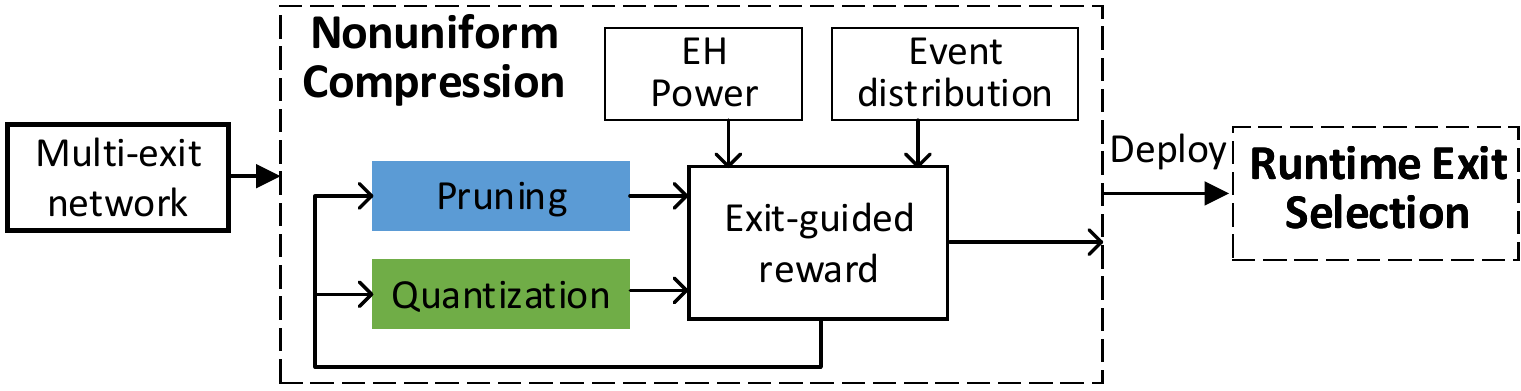}
	\caption{Overview of compression and runtime exit selection.}
	\label{fig:overview}
\end{figure}

The overview of the compression approach is shown in Figure \ref{fig:overview}.
This approach takes the multi-exit network, EH power trace, and event distribution as the input and generates non-uniform pruning rate and the bitwidth allocation policy for each layer. Based on the pruning rate, channel pruning is applied to each layer to prune out the input channels \cite{li2016pruning}. The channel to be pruned out is selected by the importance of the channel, i.e. the magnitude of weights applied to the input channel, and the less important ones are pruned out. Based on the bitwidth policy, linear quantization \cite{wang2019haq} is applied to both the weights and activations.
After compression, the network is deployed onto MCUs and the runtime algorithm will select the exit for each event, which will be introduced in Section \ref{sec:runtime}.
During compression, the approach first generates an initial layer-wise compression policy.
The compression policy prioritizes the exits with higher probability and provides them with relatively higher accuracy by adjusting the layer-wise compression policy. After applying the compression policy, the probability distribution of each exit is changed and the compression policy needs further fine-tuning. To accelerate the above iterative design process, we propose a reinforcement learning (RL)-based algorithm to co-explore the pruning and quantization policies and the probability distribution of each exit.

\subsection{Problem Formulation}

Given a full-precision network with multiple early-exits, 
we will explore the accuracy and energy cost allocation for each exit to maximize the average accuracy (equivalent to maximizing \emph{IEpmJ} defined in Section \ref{sec:inferencemodel}) under the given EH power trace and event distribution. This is achieved by non-uniformly allocating the pruning rate and quantization bitwidth for each layer.
Both pruning and quantization reduce the FLOPs and weight size of the network but with different emphasis. 
Pruning mainly reduces the FLOPs, while quantization mainly reduces the model size. 

\textbf{Pruning}
Gvien a pruning rate $\alpha_l$, we employ channel pruning to prune out the entire input channels of a convolutional or fully-connected layer. The advantages are two-fold. It reduces the FLOPs of the previous layer by reducing the number of output channels. It also reduces the FLOPs of the current layer by reducing the number of input channels. Besides, it can be directly implemented on off-the-shelf MCUs without overhead. 
More specifically, given the pruning rate $\alpha_l$ for layer $l$, we reduce the filter weights from shape $[n, c, k, k]$ to $[n, c', k, k]$ such that $\alpha_l={c'}/{c}$. For convolutional layers, $n$ and $c$ are the number of output and input channels, respectively, and $k$ is the filter kernel size. For fully-connected layers, $n$ and $c$ are the number of output and input activations, and $k=1$. The input channels to be pruned are selected according to the sum of absolute weights applied to them. We use $w_{i,j}$ to represent the weights of filter $i$ connected to input channel $j$. The importance of input channel $j$ is:
\setlength{\abovedisplayskip}{0pt}
\setlength{\belowdisplayskip}{0pt}
\setlength{\abovedisplayshortskip}{0pt}
\setlength{\belowdisplayshortskip}{0pt}
\begin{equation}
s_j=\sum_{i=1}^{n} |W_{i,j}|,\ j\in \{1,...,c\}
\end{equation}
All the input channels are sorted by $s_j$ and the least important ones are pruned out to make $c'=c$.

\textbf{Quantization}
For each layer $l$, we employ linear quantization for both the weights and activations following the bitwidth $b_l^w$ and $b_l^a$.  
Given weight bitwidth $b_l^w=k$, the linearly quantized weight $w'_l$ is:
\setlength{\abovedisplayskip}{0pt}
\setlength{\belowdisplayskip}{0pt}
\setlength{\abovedisplayshortskip}{0pt}
\setlength{\belowdisplayshortskip}{0pt}
\begin{equation}
	w'_l = clamp(round(w_l/s), -2^{k-1}, 2^{k-1}-1)\times s
\end{equation}

\noindent where $clamp(x,lb,ub)$ truncates the value $x$ into the range $[lb,ub]$ that $k$ bits can represent. $s$ is the scaling factor, which is determined by minimizing the quantization error $||w'_l-w_l||_2$.
As for activations, the quantization procedure is similar except the range for $clamp()$ is changed. Since all the activations are non-negative due to the ReLU function, we truncate the activations into the range $[0, 2^{k}-1]$.

The goal here is to find the best pruning and quatization rate. Formally, the multi-exit network compression problem under the power trace and event distribution constraints is formulated as:
\begin{align}
\text{Max} \ &\frac{1}{N}\sum_{j=1}^{N} \ Acc_{exit(j)} \label{eq:objcompression}
\end{align}
\begin{align}
\text{s.t.} \ & \sum_{j=1}^{n}\ EH_{j} \ge \sum_{j=1}^{n} E_{exit(j)}, \forall n \in \{1...N\}  \label{eq:totalEconstraint} \\
\quad & Acc_{i} = f_{acc}(\alpha_1,b_1^w,b_1^a,...,\alpha_{L_i},b_{L_i}^w,b_{L_i}^a), \quad \forall i \in \{1...m\} \label{eq:accconstraint} \\
\quad & E_{i} = f_{E}(\alpha_1,b_1^w,b_1^a,...,\alpha_{L_i},b_{L_i}^w,b_{L_i}^a), \quad \forall i \in \{1...m\} \label{eq:exitEconstraint} \\
\quad & \ S_{model} \le S_{target}, F_{model} \le F_{target} \label{eq:modelsizeconstraint} 
\end{align}

The objective is to maximize the average accuracy (equivalent to maximizing \emph{IEpmJ} defined in Section \ref{sec:inferencemodel}) of the given $N$ events and under the power trace.
In the objective function Eq.(\ref{eq:objcompression}), $Acc_{exit(j)}$ represents the accuracy of the exit for event $j$. For event $j$, an exit $i$ is selected from $m$ exits by the policy $i=exit(j)$. A simple policy is selecting the exit for an event such that the energy cost at the selected exit does not exceed currently available energy.
The first constraint listed in Eq.(\ref{eq:totalEconstraint}) is that for each of the $N$ events, the total harvested energy from the beginning to current time is greater than or equal to the total energy cost for all the happened events. Here, $EH_j$ is the harvested energy after event $j-1$ and before event $j$, and $E_{exit(j)}$ is the energy cost when exiting from exit $i$ following policy $i=exit(j)$.
The second constraint listed in Eq.(\ref{eq:accconstraint}) is that the accuracy $Acc_{i}$ of exit $i$ is determined by the pruning rate $\alpha_l$, weight bitwidth $b_l^w$ and activation bitwidth $b_l^a$ of all layers before the layer $L_i$ where exit $i$ is located. Similarly, the third constraint listed in Eq.(\ref{eq:exitEconstraint}) is that the energy cost $E_{i}$ exiting from exit $i$ is determined by all the pruning rates and bitwidth allocations before this exit. 
The last constraint listed in Eq.(\ref{eq:modelsizeconstraint}) is the weight size $S_{model}$ can fit into the target MCU and the total FLOPs $F_{model}$ is reduced to the target value $F_{target}$.

Given the pruning rate $\alpha_l$ and bitwidth $b_l^w,b_l^a, l \in \{1,...,L\}$, the objective function can be immediately calculated. This is done by first evaluating Eq.(\ref{eq:accconstraint}) on the representative dataset to get $Acc_{i}$ and measuring $E_{i}$ on the hardware or from the proxy FLOPs. 
Following the energy constraint Eq.(\ref{eq:totalEconstraint}) and exit selection policy, the exit $i=exit(j)$ for event $j\in \{1...N\}$ is determined. Given $exit(j)$, the objective function Eq.(\ref{eq:objcompression}) is calculated.
However, the search space is prohibitively large to find the optimal allocation policy. Assume the network has $L$ layers. The quantization bitwidth $b_l^w$ and $b_l^a$ are both selected from $\{1,...,8\}$, and the pruning rate $\alpha_i$ is in the range [0.05,1.0] with a step size 0.05. The design space as large as $(8^2\times20)^L \approx 10^{3L}$, which prohibits direct searching.

\subsection{RL-Based Nonuniform Compression}

\begin{figure}[!htb]
	\centering
	\includegraphics[width=\columnwidth]{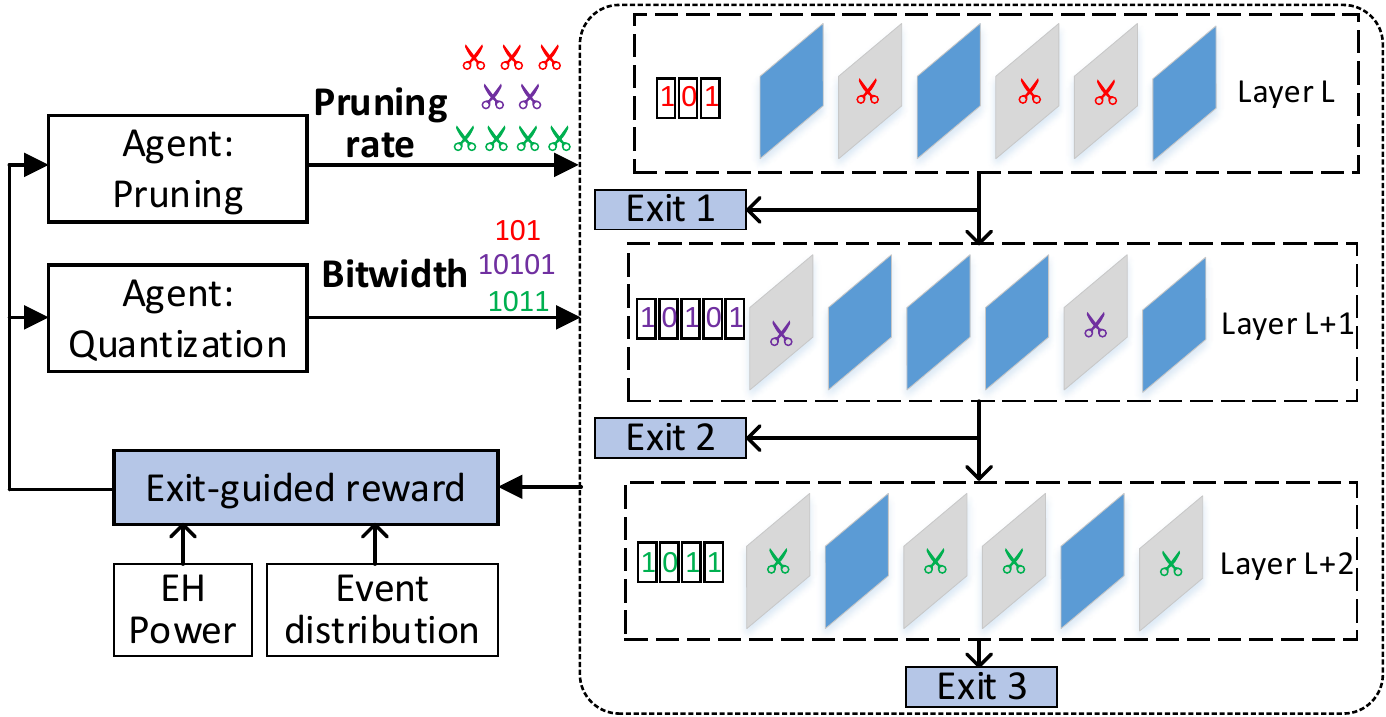}
	\caption{Exit-guided layer-wise pruning and quantization.}
	\label{fig:compressoverview}
\end{figure}
To effectively search for the optimal parameters, we model the pruning and quantization task as a reinforcement learning problem. As shown in Figure~\ref{fig:compressoverview}, we use two agents to generate the pruning rate and quantization bitwidth layer-by-layer. The compressed network is then evaluated with the EH power trace and event distribution. Here, the exit is selected according to the available energy when an event happens. After that, the reward representing the average accuracy of all events is given to the agents to update their policies.
After the exploration, the agents will generate the pruning rate and quantization bitwidth for each layer to maximize Eq.(\ref{eq:objcompression}) and equivalently \emph{IEpmJ}.

\textbf{State} Two agents share the layer-wise state during training and generate different actions. The key point is that both the pruning and quantization information are encoded in the observation. Each agent observes the peer's action in the last layer such that it can take action accordingly. For layer $l$, the shared observation $O_l$ is:
\setlength{\abovedisplayskip}{0pt}
\setlength{\belowdisplayskip}{0pt}
\setlength{\abovedisplayshortskip}{0pt}
\setlength{\belowdisplayshortskip}{0pt}
\begin{multline}
O_l = (l, \alpha_{l-1}, b_{l-1}^w, b_{l-1}^a, flop_{reduced}, flop_{remain}, \\ s_{reduced}, s_{remain}, i_{conv}, c_{in}, c_{out}, s_{weight})
\end{multline}

$l$ is the layer index. $\alpha_{l-1}$ is the pruning rate of the previous layer. $b_{l-1}^w$ and $b_{l-1}^a$ are the bitwidth of weights and activations of the previous layer. $flop_{reduced}$ is the reduced FLOPs in previous layers, and  $flop_{remain}$ is the FLOPs in the following layers. $s_{reduced}$ and $s_{remain}$ are the reduced weight size and the remaining weight size. $i_{conv}$ is a binary value indicating whether this layer is a convolutional or fully-connected layer. $c_{in}$ and $c_{out}$ are the number of input and output channels for the convolutional layer, or the number of input and output activations for the fully-connected layer. Each dimension of $O_l$ is normalized to $[0,1]$ to make them on the same scale.

\textbf{Action} Two agents generate different actions. One agent generates the action $\alpha_l$ for the layer-wise pruning rate. The other agent generates two actions, one for the layer-wise weight bitwidth $b_l^w$ and one for activation bitwidth $b_l^a$. We use continuous action space to generate accuracy pruning rate and quantization bitwidth. We do not use discrete action space because fine-grained pruning rate and quantization bitwidth need a large number of discrete actions to represent, which results in inefficient exploration during training. To apply the agents' actions to the compression process, 
the continuous action representing the pruning rate can be directly applied to pruning. The action for quantization is first linearly mapped from the continuous action space $[0,1]$ to the discrete bitwidth in the range $[b_{min}^w, b_{max}^w]$ for weights and $[b_{min}^a, b_{max}^a]$ for activations. Then the bitwidth is applied
to the quantization of weights and activations.

\textbf{Reward} 
Two agents have different reward functions $R_{prune}$ and $R_{quant}$ due to different goals. Their rewards consist of the accuracy part $R_{acc}$ and the compression part. 
$R_{acc}$ aims to maximize the average accuracy of all events under the given power trace and event distribution. 
We use the percentage of each exit being selected to guide the compression process. 
$R_{acc}$ is defined as:
\begin{equation}\label{eq:accreward}
	R_{acc}=\sum_{i=1}^m p_i Acc_i
\end{equation}

\noindent where $p_i$ is the percentage of exit $i$ being selected. It is determined by both the power trace and event distribution in Eq.(\ref{eq:objcompression})-(\ref{eq:modelsizeconstraint}). 

The compression goal of the pruning agent is to keep the FLOPs of all exits $F_{model}=\sum_{i=1}^{m}flop_i$ under the targeted value $F_{target}$. The quantization agent aims to keep the weight size $S_{model}$ under the target value $S_{target}$. Considering the accuracy reward in Eq.(\ref{eq:accreward}), the reward for two agents are defined as follows:
\begin{equation}
R_{prune} = 
\begin{cases}
\lambda_1 R_{acc} &\text{if $F_{model}\le F_{target}$}\\
-\lambda_1 &\text{otherwise}
\end{cases}
\end{equation}

\begin{equation}
R_{quant} = 
\begin{cases}
\lambda_2 R_{acc} &\text{if $S_{model}\le S_{target}$}\\
-\lambda_2 &\text{otherwise}
\end{cases}
\end{equation}

\noindent where $\lambda_1$ and $\lambda_2$ are the reward scaling factors. When the compression goal is satisfied, the reward is the scaled accuracy. Otherwise, the reward is a negative value to punish the agents.

\textbf{Agent} 
We use two RL agents, one for pruning and the other for quantization.  Separate agents enable us to set different rewards to achieve different goals simultaneously. The agents leverage the deep deterministic policy gradient (DDPG)~\cite{lillicrap2015continuous} algorithm to explore the design space. The agents process the network layer-by-layer. In the learning process, one step represents the agent processes one layer. For each layer, two agents take the step simultaneously and proceed to the next layer. One episode consists of many steps. It starts from the first layer and ends at the last layer.

During exploration, each agent aims to maximize the overall reward of one episode. The action-value Q-function is estimated as 
\begin{equation}
Q_l^\prime = r_l + Q(O_{l+1}, a_{l+1}) |_{a_{l+1}=\mu (O_{l+1})}
\end{equation}

The Q-function $Q(O,a)$ is updated by minimizing the loss:
\begin{equation}
	Loss=\frac{1}{N}\sum_{l}(Q_l^\prime - Q(O_{l}, a_{l})) |_{a_{l}=\mu (O_{l})}
\end{equation}

\noindent where $N$ is the number of sampled steps during exploration. The policy $a=\mu(O)$ is updated using the sampled policy gradient:
\begin{equation}
\nabla J = \frac{1}{N}\sum_{l} \nabla_{a_l} Q(O_l,a_l) \nabla \mu (O_l)
\end{equation}

\section{Runtime Exit Selection and Incremental Inference}\label{sec:runtime}

During the compression process, the exit selection for an event $j$ is determined statically using a static policy, e.g. a lookup table (LUT). However, naively following the static policy during runtime can result in low average accuracy in the long term. For example, when the EH power is low in the long run, even if the system has sufficient energy to select the exit with the highest accuracy and energy cost for the current inference, a better decision can be selecting an exit with lower energy cost to reserve energy for following events. This dynamic exit-selection can improve the average accuracy. 
Besides, if the confidence at the selected exit is low, an incremental inference by proceeding to the following exit can improve the accuracy.  
We propose an online algorithm to make these two sequential decisions. 

During runtime, both the power trace and the event distribution are unknown in advance. 
To select the best exit for each event, we propose to employ a lightweight RL algorithm, Q-learning \cite{watkins1992q}.
Q-learning consists of the state set $\mathcal{S}$, the action set $\mathcal{A}$ and the reward function $R$. 
The state set $\mathcal{S}$ contains the current available energy $E$ and the charging efficiency $P$. Since both $E$ and $P$ are continuous values, to make the number of elements in $\mathcal{S}$ finite, we discretize $E$ and $P$ with appropriate step size. 
The action set $\mathcal{A}$ represents all the possible exits, which is $\mathcal{A}=\{exit_1,...,exit_m\}$. 
The reward $R$ is the accuracy of the selected exit $r=Acc_{a}, a\in \mathcal{A}$. 
The agent aims to learn the optimal policy $\pi$ such that $a=\pi(s),a\in \mathcal{A}, s\in \mathcal{S}$ to maximize the reward $R=\sum r$. When an event happens, the agent takes two steps, one for selecting the action and the other for updating the Q-table. The action for the exit is selected by finding the highest Q-value in current state, represented as $a=\arg \max_{a \in \mathcal{A}} Q(s, a)$, where
$Q(s,a)$ denotes the Q-value of action-state pair $(s,a)$. 
The entry $(s, a)$ in the Q-table is updated as:
\begin{equation}\label{eq:qlearning}
	Q(s, a) = Q(s, a)+\alpha(r+\gamma \max_{a \in \mathcal{A}} Q(s', a)-Q(s, a))
\end{equation}
The overhead of Q-learning is negligible. It only needs a lookup table (LUT) with state-action pairs as the entries, and the learning process is updating the LUT by Eq.(\ref{eq:qlearning}).

To further improve the average accuracy, a second decision is made at the chosen exit for event $j$. If the confidence of the result is low and the remaining energy is high, the algorithm can decide to propagate the input further to the next exit for higher accuracy. 
The decision is made based on the confidence of the result and current available energy. We use the entropy of the result as the measure of confidence~\cite{teerapittayanon2016branchynet}. We use another Q-table to make the decision.

\section{Experimental Results}\label{sec:experiment}

We conduct extensive experiments to demonstrate the effectiveness of our approaches in terms of \emph{nonuniform compression}, \emph{IEpmJ and accuracy}, \emph{FLOPs and latency},  and \emph{runtime adaptation}.

\subsection{Experimental Setup}
The experiments are targeting on TI MSP432 MCU. To power the MCU, we use solar profile from \cite{ornl2007}. 
The backbone of the multi-exit model is LeNet \cite{lecun1998gradient}. 
We use LeNet because most state-of-the-art DNNs designed for mobile devices cannot fit into typical MCUs even after compression. For example, MobileNetV2~\cite{sandler2018mobilenetv2} and DARTS \cite{liu2018darts}
require 4.6MB and 6.6MB weight storage, respectively. However, a typical MCU has tens of KBs weight storage.
We extend LeNet to four convolutional layers and equip it with two early-exits along the data path. 
The original network needs 580KB weight storage when represented with 32-bit floating-point numbers. The FLOPs of three exits are 0.4452M, 1.2602M and 1.6202M with corresponding accuracy 64.9\%, 72.0\% and 73.0\%.
The energy cost is 1.5mJ per million FLOPs.
We are using the CIFAR-10 dataset and 500 events are randomly distributed across the duration of the EH power trace.

\subsection{Nonuniform Pruning and Quantization}
\begin{figure}[!htb]
	\centering
	\includegraphics[width=\columnwidth]{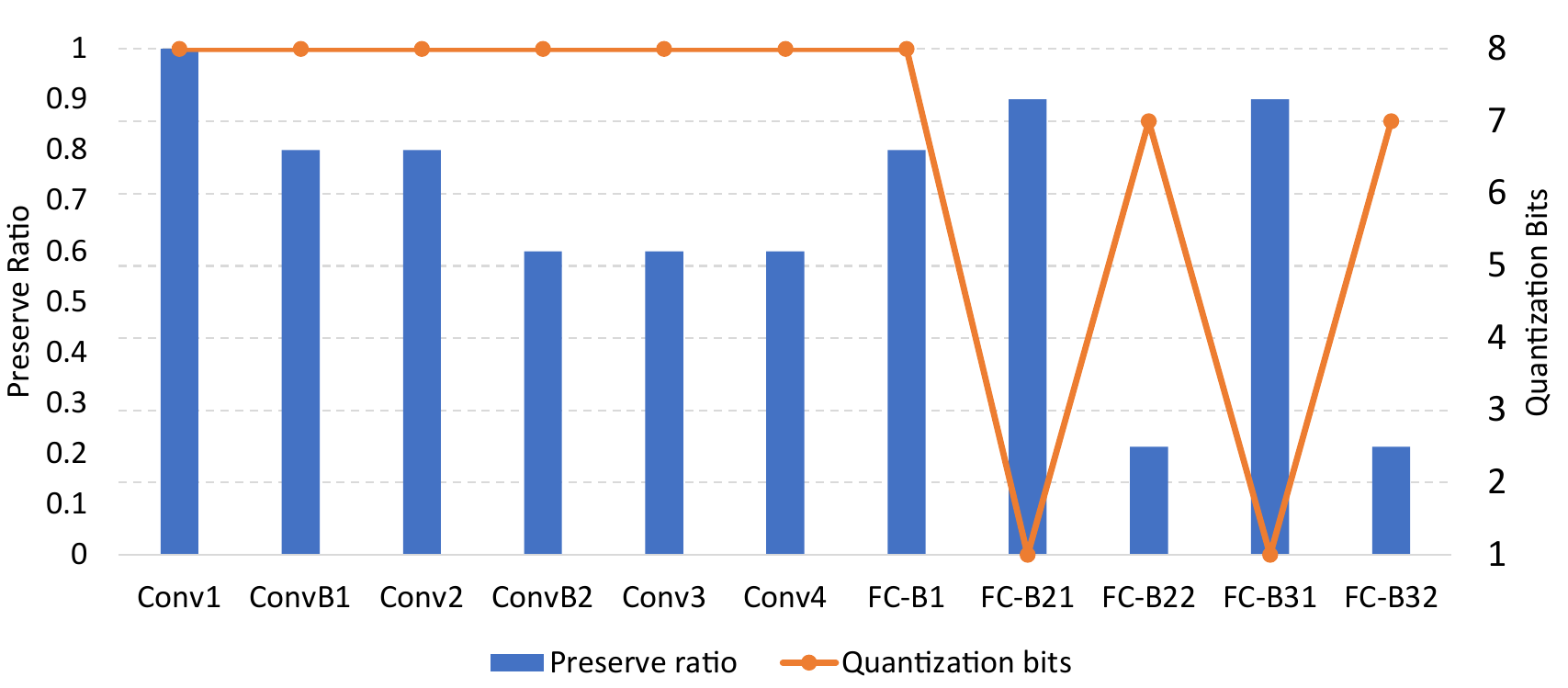}
	\caption{Pruning and quantization policy under 1.15M FLOPs and 16KB weight size constraints.}
	\label{fig:nonuniform}
\end{figure}

Our approach effectively finds out the pruning rate and quantization bitwidth allocation policy to maximize the average accuracy under the model size and FLOPs constraint. Figure~\ref{fig:nonuniform} shows the layer-wise preserve rate and quantization bitwidth. 
The FLOPs constraint is set to 1.15M FLOPs, and the target model size is set to 16KB.
Under these constraints, our approach efficiently allocates the limited FLOPs and weight size budge to maximize the accuracy. For pruning, the convolutional layers are pruned more because they are more FLOPs-intensive than the fully-connected layers. 
Different from pruning, the quantization allocates more accuracy to convolutional layers by setting their bitwidth to 8.
FC-B21 and FC-B31 are quantized to 1-bit possibly because they have large weight size and less sensitive to data precision. 
The search takes 6 hours on a Nvidia P100 GPU.
\subsection{IEpmJ and Average Accuracy}
\begin{figure}[!htb]
	\centering
	\includegraphics[width=0.8\columnwidth]{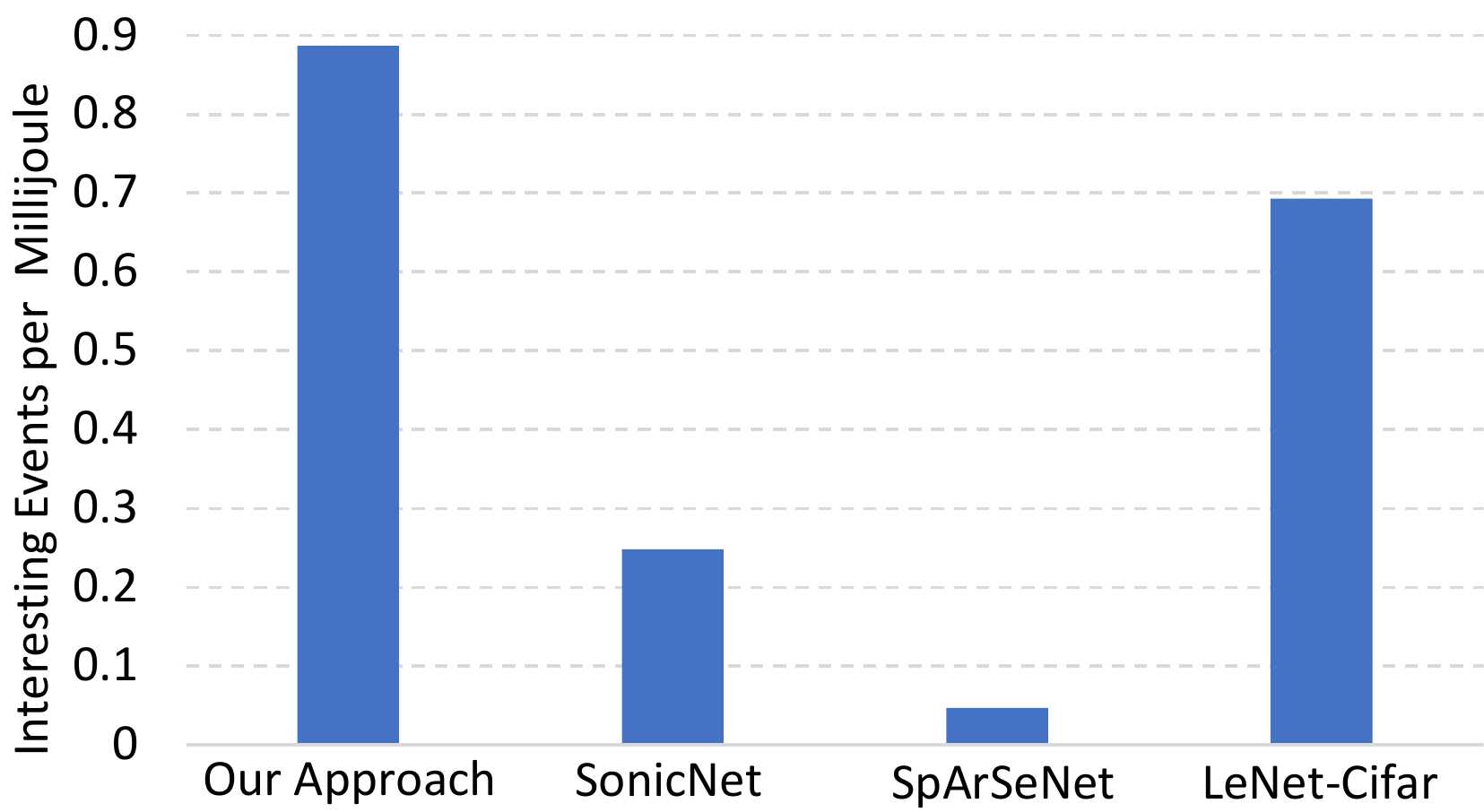}
	\caption{Number of interesting events per energy harvesting millijoule.}
	\label{fig:IEmpJ}
\end{figure}

The proposed approaches substantially outperform the SOTA baselines in terms of \emph{IEpmJ} (Interesting Events per milliJoule) and equivalently the average accuracy of all events. 
We compare with three baselines. SonicNet is from the SOTA intermittent inference framework~\cite{gobieski2019intelligence}.
SpArSeNet is a network generated by a Neural Architecture Search framework for MCUs~\cite{fedorov2019sparse}. LeNet-Cifar is the LeNet~\cite{lecun1998gradient} adapted for CIFAR-10 dataset. 

The result of \emph{IEpmJ} is shown in Figure \ref{fig:IEmpJ}. 
Our approach outperforms SonicNet, SpArSeNet and LeNet-Cifar by 3.6x, 18.9x and 0.28x, respectively.
Our approach achieves 0.89 interesting events per milljoule, while SonicNet and SpArSeNet only achieve 0.25 and 0.05, respectively.  
During compression, our approach considers the accuracy and energy cost of each exit, the EH power trace and the event distribution to compress the network such that the \emph{IEpmJ} is maximized. 
In terms of the accuracy of all events, where the accuracy of missed event is set to 0, our approach achieves average accuracy of 50.1\%, while SonicNet, SpArSeNet and LeNet-Cifar only achieve 14.0\%, 2.6\% and 39.2\%, respectively. 
As for the accuracy of all the processed events, our approach achieves 65.4\%, slightly lower than 75.4\%, 82.7\%, 74.7\% by the baselines. This is because we aim to improve the long-term accuracy to maximize \emph{IEpmJ} instead of the accuracy for a single event. Solely aiming at the per-inference accuracy will generate network with high energy cost and result in high percentage of missed events, which degrades \emph{IEpmJ}.

\subsection{FLOPs and Latency}
\begin{figure}[!htb]
	\centering
	\includegraphics[width=0.9\columnwidth]{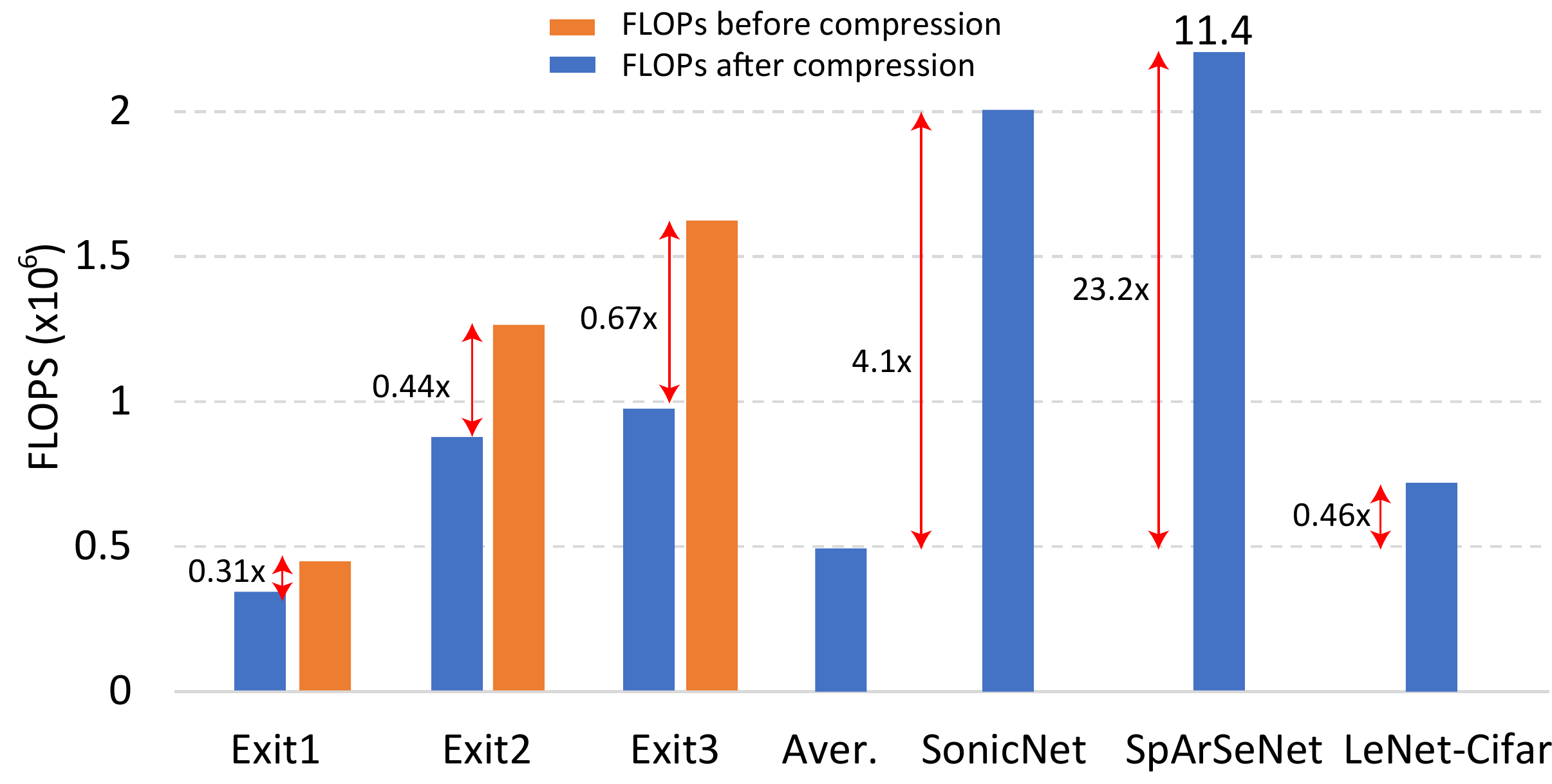}
	\caption{FLOPs reduced by compression}
	\label{fig:FLOPsreduce}
\end{figure}

\textbf{FLOPs} Our approach effectively reduces the FLOPs of each exit to maximize the average accuracy of all events. 
Reducing FLOPs is important because with lower FLOPs and lower energy cost per inference, the saved energy can be allocated to other events which could have been missed due to insufficient energy.
Figure~\ref{fig:FLOPsreduce} shows the FLOPs of each exit before and after compression. The FLOPs are reduced by 0.31x, 0.44x and 0.67x for three exits, respectively. 
The reduction ratio of each exit is automatically decided by our approach. 
Different from our approach, the SonicNet has 2.0M FLOPs and SpArSeNet has 11.4M FLOPs because they did not consider the limited EH energy and only prioritize the per-inference accuracy. This results in high energy cost per inference, low \emph{IEpmJ} and low average accuracy across all events because large portion of the events are missed. The LeNet-Cifar is manually designed by domain experts and has low FLOPs, which fortunately fits the EH scenario well. 

\textbf{Latency} Our approach greatly reduces both \emph{per-event} latency and \emph{per-inference} latency. First, the \emph{per-event} latency is from the occurrence of an event to the end of inference. Across all the processed events, our approach improves the per-event latency by 7.8x, 10.2x and 3.15x over three baselines. More specifically, the average latency of our approach is 18.0 time units (1 second per time unit), while the latency of three baselines are 139.9, 183.4 and 56.7 time units, respectively. The improvement shows our approach smartly selects the early-exits to quickly output a result when the EH energy is low, instead of waiting for multiple power cycles to reach the final exit as the baselines do.
Second, our approach also improves the \emph{per-inference} latency, which is from the start to the end of an inference.
As shown in Figure~\ref{fig:FLOPsreduce}, using the FLOPs as the proxy for the \emph{per-inference} latency, our approach improves the average \emph{per-inference} latency by 4.1x, 23.2x and 0.46x over three baselines.

\subsection{Runtime Adaptation}
\begin{figure}[!htb]
	\centering
	\begin{subfigure}{0.49\columnwidth}
		\includegraphics[width=\columnwidth]{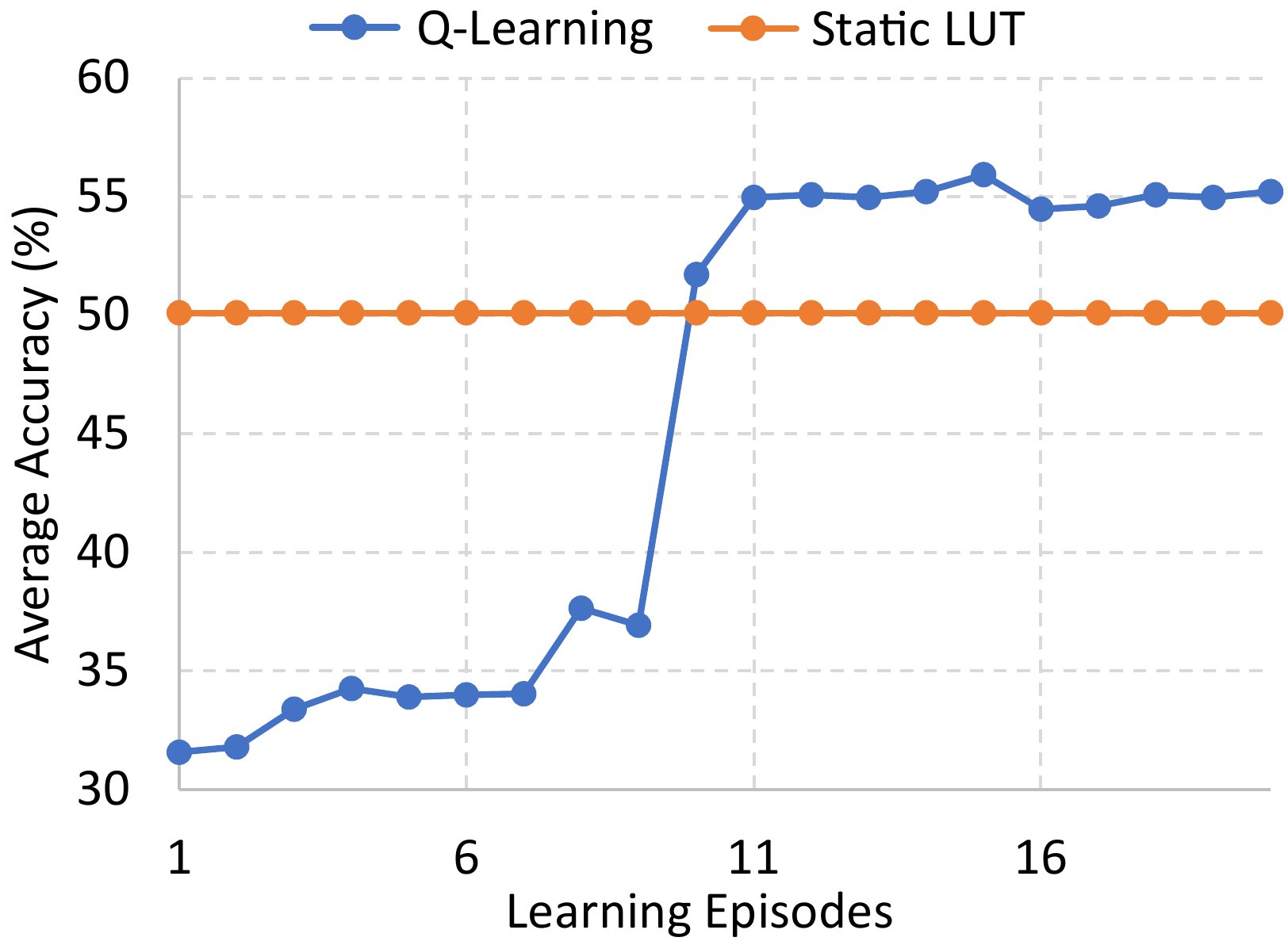}
		\caption{Runtime learning process.}
		\label{fig:q_learning_process}
	\end{subfigure}
	\begin{subfigure}{0.49\columnwidth}
		\includegraphics[width=\columnwidth]{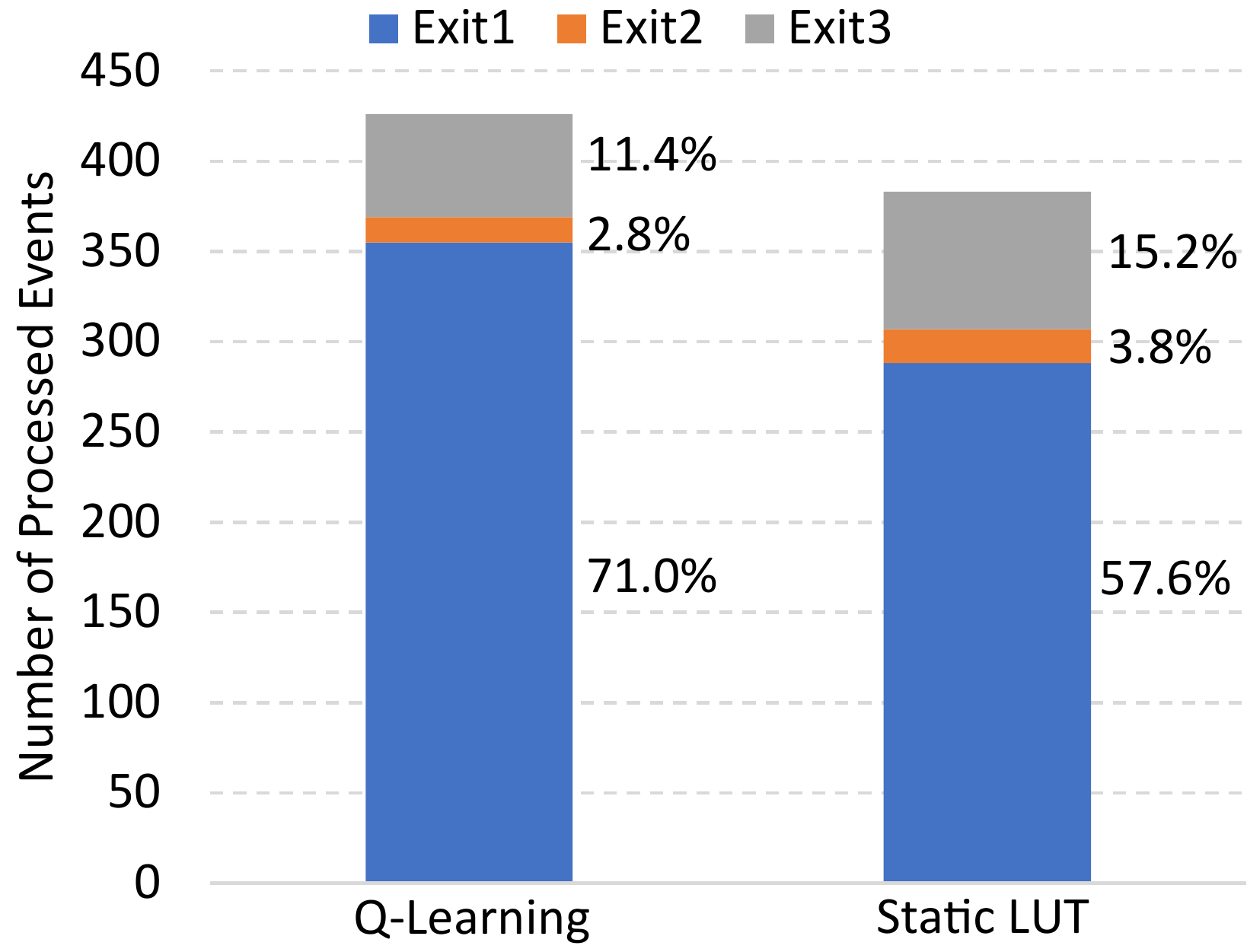}
		\caption{Number of processed events.}
		\label{fig:q_events}
	\end{subfigure}
	\caption{Runtime adaptation by lightweight learning.}
	\label{fig:runtime}
\end{figure}

The average accuracy of all events is further improved by the runtime exit selection. The runtime adaptation effectively learns from the EH environment and selects the exit for each event to maximize the average accuracy. The adaptation approach outperforms the static LUT by 10.2\%.
Figure~\ref{fig:q_learning_process} shows the average accuracy of all events is improved during the runtime adaptation. The lightweight Q-learning approach gradually learns to optimize the exit selection. 
Figure~\ref{fig:q_events} shows the percentage and number of events exiting from each of the three exits. Compared with the static LUT, the Q-learning approach prioritizes the exit 1 shown in the blue bar to decrease the energy cost of each inference. 
By the strategy adaptation, the Q-learning approach processes 11.2\% more events than the static LUT. The overhead of Q-learning is negligible by updating its Q-table.

\section{Related Work} \label{sec:related}
\noindent\textbf{Intermittent Execution.} 
EH techniques extract power from the ambient environment and provide an attractive power alternative in sensing scenarios
where it is difficult to employ batteries \cite{wu2019work}. 
Solar, wind and kinetic energy \cite{ju2018power} are all promising EH sources. 
With an unstable power supply, EH-powered computing systems have to run intermittently \cite{jia2019q}. 
Various optimization and tools such as Chain~\cite{colin2016chain} have also been proposed to ensure correctness and improve efficiency.
Gobieski et al. \cite{gobieski2019intelligence} made the first step to implement DNNs in an intermittently-powered sensor.
It guarantees the correctness of DNN inference across multiple power failures. 
The drawback is that we must wait for multiple power cycles to finish one inference. Since the harvested power is usually weak and unpredictable, it takes indefinite amount of time to obtain the final inference result.

\noindent\textbf{Multi-Exit Network.}
The multi-exit neural network has been investigated in various studies. Instead of only having one final inference result, networks can have early result to save time or energy.
In \cite{teerapittayanon2016branchynet,huang2017multi}, a subset of networks is selected for faster inference by trading off accuracy.
These approaches allow dynamic trade-off between inference latency and accuracy. 
However, none of the works are targeted on EH-powered MCUs, which are constrained in the weight storage and energy budgets. The large weight size and FLOPs of their models are prohibitive for direct deployment to EH-powered MCUs. Pruning and quantization are needed for the deployment. 

\noindent\textbf{Network Compression.}
There are extensive explorations on network pruning and quantization.
For quantization, 
\cite{rastegari2016xnor} employs binary filters and inputs for CNNs.
\cite{wang2019haq} automates the quantization of each layer by a learning-based method. 
\cite{lu2019neural,bian2020nass} consider quantization during the neural architecture search (NAS) for efficient hardware implementation \cite{jiang2019accuracy}.
For pruning, 
\cite{he2018amc} employs RL to automatically explore the layer-wise pruning rate for channel pruning \cite{li2016pruning}. 
However, these pruning and quantization methods only consider network with one exit, which will greatly degrade the accuracy of early-exits.
Besides, the above approaches only focus on either quantization or pruning, not both of them. To deploy multi-exit networks to MCUs, an automated approach to conduct the quantization and pruning simultaneously while considering the accuracy of all exits is needed.

\section{Conclusion}
This work aims to enable event-driven sensing and decision capabilities for EH-powered devices by deploying lightweight DNNs onto EH-powered devices. 
We provide an intermittent inference model to provide timely and accuracy result. We propose a two-phase approach to deploy multi-exit neural networks onto EH-powered MCUs.
For the first phase, we develop a power trace-aware and exit-guided network compression algorithm to compress the networks to maximize the overage accuracy. For the second phase, we develop a runtime exit selection algorithm to adapt to dynamic EH environment and event distribution. The experimental results show superior accuracy and latency compared with state-of-the-art techniques.

\vspace{3pt}
\noindent
\textbf{Acknowledgement: }
This work used the Extreme Science and Engineering Discovery Environment (XSEDE), which is supported by National Science Foundation grant number ACI-1548562. Specifically, it used the Bridges system, which is supported by NSF award number ACI-1445606, at the Pittsburgh Supercomputing Center (PSC).
This research was supported in part by the University of Pittsburgh Center for Research Computing through the resources provided.

\bibliographystyle{IEEEtran}
\bibliography{IEEEexample}

\end{document}